\documentclass[twoside,11pt]{article}

\usepackage[abbrvbib, preprint]{jmlr2e}

\usepackage{bussproofs}
\usepackage{amsmath}
\usepackage{algorithm}
\usepackage[noend]{algpseudocode}

\usepackage{lastpage}

\begin{document}

\title{Greedy Grammar Induction with Indirect Negative Evidence}

\author{\name Joseph Potashnik \email joseph.potashnik@gmail.com \\
       \addr London, United Kingdom
}

\maketitle
\begin{abstract}
This paper proposes a non-lexicalized grammar-induction procedure that separates
two tests: recognition of the observed finite presentation, and rejection of
short preterminal strings generated by a hypothesis but unsupported by the
evidence.

The central object is the rule-coverage bound \(\ell^*(G)\): the maximum, over
rules in \(G\), of the length of the shortest preterminal string whose
derivation uses that rule. This bound induces the comparison universe
\(\Sigma_{\mathrm{pre}}^{\le \ell^*(G)}\), where unsupported generated strings
serve as indirect evidence against overgenerating hypotheses.

We give a greedy search algorithm over rule sets and prove a
conditional weak-recovery theorem: under explicit reachability conditions and
sufficient saturation of the presentation, the exact learner reaches a grammar
weakly equivalent to the unknown target. The complexity analysis is slice-wise: for each fixed incrementality
radius \(k\), the search explores polynomially many rule-set extensions
in the finite rule universe. Across 31 benchmark runs spanning Dyck-\(k\)
languages \((1\le k\le4)\), palindromes, \(a^n b^n\), English-like recursive
fragments, and an inherently ambiguous union language, grammar-level analysis
establishes weak equivalence between every returned grammar and its target.
\end{abstract}

\begin{keywords}
context-free grammar induction, grammatical inference, learning from positive data, indirect negative evidence, rule-coverage bounds
\end{keywords}

\section{Introduction}
\label{sec:Introduction}

Context-free grammars are natural targets for grammar induction because they
give compact symbolic descriptions of recursive structure. Learning them from positive evidence alone, however, raises two separate
problems. The first is criterion design: under what assumptions can positive
strings distinguish a target grammar from larger grammars that also contain the
observed sample? The second is search: even when a criterion is fixed, the space
of rule sets and latent derivations is combinatorial.

Classical identification results make the first difficulty precise: rich
language classes cannot be identified from arbitrary positive texts alone
\citep{gold1967language,angluin1980inductive}. Positive results in grammatical
inference therefore typically restrict the target class, require characteristic
samples, change the representation, or enrich the examples with structural
information \citep{delahiguera1997characteristic,
delahiguera2010grammatical,clark2007polynomial,clark2010contextual,
sakakibara1992efficient}. A different line of work places a probabilistic model
over strings and latent derivations and optimizes a likelihood or variational
objective, as in EM-trained grammar models and neural PCFGs
\citep{dempster1977maximum,klein2002generative,klein2004corpus,
kim-etal-2019-compound}. These methods have been empirically useful, but the cited systems are optimized
and evaluated primarily by likelihood or induced-structure accuracy, rather than
by a guarantee that the returned grammar is weakly equivalent to an unknown
target. Their objectives also lead to non-convex latent-variable search
problems.

This paper takes a symbolic route. The learner is non-lexicalized: it induces a
context-free grammar over preterminal strings, assuming an external
word-to-preterminal lexicon. This separation is also common in neural PCFG work,
where nonterminals, preterminals, and terminals are represented as distinct
grammar components \citep{kim-etal-2019-compound,yang-etal-2021-pcfgs}. A
candidate grammar must satisfy two tests. It must recognize the observed finite
presentation, and it must avoid generating short preterminal strings
unsupported by the evidence. Thus positive data are used not only as examples
to cover, but also as indirect evidence against bounded overgeneration.

The key object is the rule-coverage bound \(\ell^*(G)\). For each rule in a
grammar \(G\), consider the shortest generated preterminal string whose
derivation uses that rule. The maximum of these witness lengths is
\(\ell^*(G)\). This bound induces the comparison universe
\(\Sigma_{\mathrm{pre}}^{\le \ell^*(G)}\): within this universe, generated
strings can be checked against the strings supported by the presentation. When
the presentation is sufficiently saturated, non-occurrence within this bounded
universe becomes informative
\citep{clark2010linguistic,hsu2016sampling}.

The fitness test does more than evaluate completed candidates: together with
rule-addition monotonicity, it shapes the rule lattice into a greedy search
landscape. Adding rules can create derivations but cannot delete them; hence,
once a candidate generates an unsupported short string, later rule additions
cannot repair that error. Under a saturated presentation, productive
subgrammars of the target remain on the fit plateau, while extensions that
generate unsupported short strings leave that plateau permanently. Moreover,
shortest unparsed evidence strings commute along target-preserving paths:
discovering one such string first does not destroy the possibility of
discovering another later. The learner exploits this structure by expanding fit
roots through bounded rule increments and by using the next shortest unparsed
string to guide candidate generation and preterminal abduction.

The result is a finite-search recovery theorem, not an
identification-in-the-limit theorem. If a grammar weakly equivalent to the
target can be reached in the hypothesis space by bounded rule increments, and
the presentation is sufficiently saturated to witness those increments, then
the exact learner is guaranteed to reach that grammar.

Outside these conditions, a returned grammar is guaranteed only to parse the
observed presentation and to avoid unsupported generated strings up to its own
rule-coverage bound, as expected from the classical limits on identification
from positive data alone.

This paper makes four contributions:
\begin{enumerate}
    \item \textbf{Rule-coverage fitness.} It defines the rule-coverage bound
    \(\ell^*(G)\) and the induced fitness test over
    \(\Sigma_{\mathrm{pre}}^{\le \ell^*(G)}\), turning unsupported short
    generated strings into evidence against overgeneration.

    \item \textbf{Fitness-induced greedy landscape.} It proves the monotonicity,
    persistence, and shortest-string commutativity properties that justify
    greedy search over fit subgrammars under saturated presentations.

    \item \textbf{Conditional weak recovery.} It gives explicit reachability
    and presentation-saturation conditions under which the exact learner reaches
    a grammar weakly equivalent to the target.

    \item \textbf{Implementation and evaluation.} It implements the learner
    using canonical adjacency, wildcard-guided generation, CKY preterminal
    abduction, Pareto-ribbon filtering, and strong-equivalence tracking. The
    complexity analysis is slice-wise: for fixed incrementality radius, rule-set
    search is polynomial in the finite rule universe, while evaluation is
    output-sensitive in the number of generated preterminal strings. Across 31
    benchmark runs, grammar-level analysis establishes weak equivalence between
    every returned grammar and its target.
\end{enumerate}

The remainder of the paper is organized as follows. Section
\ref{sec:sufficient_texts} defines witness strings and the rule-coverage bound.
Section \ref{sec:Fitness Function} introduces the fitness function. Section
\ref{sec:Learning Algorithm} presents the learner. Sections
\ref{sec:Complexity} and \ref{sec:Learnability} analyze complexity and
conditional recovery. Section \ref{sec:SampleComplexity} gives sample
requirements for saturation. Section \ref{sec:EmpiricalEvaluation} reports the
experiments, Section \ref{sec:future} discusses limitations and future work,
and Section \ref{sec:Conclusion} concludes.

\section{Sufficient Texts and Rule Coverage Bound}
\label{sec:sufficient_texts}

This section defines the grammar-internal length scale used later for the
fitness test.

\subsection{Finite-Horizon Sufficiency}

We begin with a grammar-internal notion of rule exposure. A finite presentation
may contain enough positive strings to witness every rule of a target grammar,
even though the learner does not know the target grammar in advance. This notion
provides the length scale used below to compare generated short strings with
the evidence.

\begin{definition}[Witness string]
Let \(G=(V,\Sigma,R,S)\) be a context-free grammar. A string
\(w\in\Sigma^*\) is a \emph{witness} for a rule \(r\in R\) if there is a
derivation
\[
S=\alpha_0 \Rightarrow \alpha_1 \Rightarrow \cdots \Rightarrow \alpha_n=w
\]
in which at least one step applies \(r\).
\end{definition}

\begin{definition}[Rule-sufficient text]
A finite presentation \(D\) is \emph{rule-sufficient} for a target grammar
\(G=(V,\Sigma,R,S)\) if every rule \(r\in R\) has at least one witness in
\(D\).
\end{definition}

\subsection{Minimal Witness Length}

\begin{definition}[Minimum derivation length]
For \(X\in V\cup\Sigma\), define
\[
\mu(X)=\min\{|w|: X\Rightarrow^*w,\; w\in\Sigma^*\}.
\]
For terminals \(a\in\Sigma\), \(\mu(a)=1\); for \(\epsilon\), \(\mu(\epsilon)=0\);
and if \(X\) is unproductive, \(\mu(X)=\infty\). For a sequence
\(\alpha=X_1\cdots X_k\), define
\[
\mu(\alpha)=\sum_{j=1}^k \mu(X_j).
\]
\end{definition}

\begin{definition}[Minimum context length]
For \(A\in V\), define
\[
c(A)=\min\{|u|+|v|:S\Rightarrow^*uAv,\; u,v\in\Sigma^*\}.
\]
Thus \(c(S)=0\), and \(c(A)=\infty\) if \(A\) is unreachable from \(S\).
\end{definition}

\begin{definition}[Minimal witness length]
For a rule \(r:A\to\alpha\), define
\[
L_r(G)=\min\{|w|:w\in L(G)\ \text{and}\ w\ \text{witnesses}\ r\}.
\]
\end{definition}

\begin{proposition}[Minimal witness length]
For every reachable and productive rule \(r:A\to\alpha\),
\[
L_r(G)=c(A)+\mu(\alpha).
\]
\end{proposition}

\begin{proof}
Any derivation witnessing \(r\) reaches a sentential form containing \(A\),
applies \(A\to\alpha\), and then derives a terminal string. Since derivations
inside distinct context-free subtrees are independent, the shortest such
terminal yield is obtained by combining a shortest terminal context for \(A\)
with a shortest terminal yield of \(\alpha\). This gives a witness of length
\(c(A)+\mu(\alpha)\). Conversely, any shorter witness would either place \(A\)
in a shorter terminal context than \(c(A)\), or derive \(\alpha\) to a shorter
terminal yield than \(\mu(\alpha)\), contradicting the definitions.
\end{proof}

\begin{definition}[Rule-coverage bound]
For a grammar \(G=(V,\Sigma,R,S)\), the \emph{rule-coverage bound} of \(G\) is
\[
\ell^*(G)=\max_{r\in R}L_r(G).
\]
It is finite exactly when every rule in \(R\) is reachable and productive. This
scalar bound induces the finite universe \(\Sigma^{\le \ell^*(G)}\), within
which every rule in \(R\) has a witness.
\end{definition}

The bound \(\ell^*(G)\) is computable in polynomial time for a fixed grammar.
The values \(\mu(X)\) are the min-sum, or Viterbi-inside, weights of the
grammar, and the values \(c(A)\) are the corresponding outside/context weights.
They can be computed by standard fixed-point and shortest-path procedures in
\(O(|V|\cdot |R|_{\mathrm{occ}})\) and
\(O(|R|_{\mathrm{occ}}+|V|\log |V|)\) time, respectively, where
\(|R|_{\mathrm{occ}}\) is the total number of right-hand-side symbol
occurrences. These are standard computations, so we use only their complexity below.

\section{The Fitness Landscape and Structural Properties}
\label{sec:Fitness Function}

This section defines the fitness score used by the learner and records the
monotonicity properties that make greedy rule-set search possible.

\subsection{The Fitness Function}

The learner uses positive evidence to test overgeneration only inside a finite
horizon. Under the saturation assumption defined below, a short string generated
by a candidate but absent from the presentation is evidence against that
candidate. The objective therefore compares the finite-horizon language
generated by a grammar with the evidence observed in the same window.

Throughout this section, \(D\) denotes the set of observed string types; token
multiplicities are ignored. For a horizon \(H\), the finite comparison universe
is \(\Sigma^{\le H}\). For the definitions and propositions in this section,
\(H\) is held fixed. When the learner evaluates a candidate grammar \(G\), this
horizon is instantiated as \(H=\ell^*(G)\).

\begin{definition}[Count vectors]
For a grammar \(G\) and horizon \(H\), define the grammar counts vector
\[
\mathbf{GC}_H(G)=(GC_0,\ldots,GC_H),\qquad
GC_i=|L(G)\cap\Sigma^i|.
\]
For a presentation \(D\), define the evidence counts vector
\[
\mathbf{EC}_H(D\mid G)=(EC_0,\ldots,EC_H),\qquad
EC_i=|D\cap L(G)\cap\Sigma^i|.
\]
\end{definition}

The grammar counts are computed by materializing the unique strings generated
by \(G\) up to length \(H\) using a standard bottom-up dynamic program over the
rule set. The procedure counts string types rather than derivations, so
multiple derivations of the same string contribute a single count. Equivalently,
the DP maintains sets \(D_A[\ell]\) of distinct length-\(\ell\) preterminal
strings derivable from each nonterminal \(A\): lexical rules initialize
length-one cells, binary rules add concatenations over all length splits, and
unit rules are closed to a fixed point. The computation is output-sensitive: in
the worst case, the number of generated strings is exponential in \(H\). The approach is therefore intended for the small universes induced by the
rule-coverage bound, not for unrestricted language enumeration.

\begin{definition}[Finite-horizon fitness]
\label{def:fitness_function}
For a grammar \(G\), presentation \(D\), and horizon \(H\) with
\(\sum_{i=0}^H GC_i>0\), define
\begin{equation}
f_H(D\mid G)=
\frac{\sum_{i=0}^{H}EC_i(D\mid G)}
     {\sum_{i=0}^{H}GC_i(G)}.
\label{fitnessfunctiondef}
\end{equation}
\end{definition}

A score of \(1\) means that every string generated by \(G\) within the horizon
is present in the evidence support. Scores below \(1\) indicate overgeneration
relative to the observed presentation within that horizon. This score is not a
full acceptance criterion: it is paired with recognition of the entire finite
presentation, including strings longer than \(\ell^*(G)\).

Unless another horizon is explicitly displayed, we use \emph{fitness} to mean
this score at the rule-coverage bound:
\[
f(D\mid G)=f_{\ell^*(G)}(D\mid G).
\]
\begin{proposition}[Persistence of finite-horizon overgeneration]
\label{prop:overgeneration_persistence}
Let \(D\subseteq\Sigma^*\) be a finite presentation and let \(G\subseteq G'\)
be grammars over the same terminal alphabet and start symbol. For any fixed
horizon \(H\), if there exists \(w\in L(G)\cap\Sigma^{\le H}\) with
\(w\notin D\), then \(w\in L(G')\cap\Sigma^{\le H}\) and \(w\notin D\).
Thus rule addition cannot remove an unsupported string already generated within
a fixed horizon.
\end{proposition}

\begin{proof}
Every derivation in \(G\) is also a derivation in \(G'\), since \(G'\) contains
all rules of \(G\). Hence \(w\in L(G')\). The presentation \(D\) is fixed, so
\(w\notin D\) remains true.
\end{proof}

\subsection{Optimal Substructure and Monotonicity}
\label{sec:Optimal_Substructure}

For a fixed horizon \(H\), let \(G=(V,\Sigma,R,S)\) and
\(G'=(V,\Sigma,R',S)\) with \(R\subseteq R'\). Let
\[
GC=\sum_{i=0}^{H}|L(G)\cap\Sigma^i|,
\qquad
EC=\sum_{i=0}^{H}|D\cap L(G)\cap\Sigma^i|,
\]
and define \(\Delta GC\) and \(\Delta EC\) analogously as the increases in
these aggregate counts from \(G\) to \(G'\).

\begin{proposition}[Marginal fitness and monotonicity]
\label{prop:marginal_monotonicity}
Assume \(GC>0\) and \(\Delta GC>0\). Then
\[
f_H(D\mid G') \ge f_H(D\mid G)
\quad\Longleftrightarrow\quad
\frac{\Delta EC}{\Delta GC} \ge f_H(D\mid G).
\]
If \(\Delta GC=0\), the bounded generated language is unchanged, and the
finite-horizon fitness is stationary.
\end{proposition}

\begin{proof}
The aggregate counts for \(G'\) are \(EC+\Delta EC\) and \(GC+\Delta GC\).
Since \(GC>0\) and \(GC+\Delta GC>0\), the inequality
\[
\frac{EC+\Delta EC}{GC+\Delta GC}\ge\frac{EC}{GC}
\]
is equivalent, by cross-multiplication, to
\[
GC\cdot\Delta EC \ge EC\cdot\Delta GC,
\]
and hence to
\[
\frac{\Delta EC}{\Delta GC}\ge\frac{EC}{GC}=f_H(D\mid G).
\]
If \(\Delta GC=0\), rule addition generates no new string types within the
fixed horizon, so \(\Delta EC=0\) and the ratio is unchanged.
\end{proof}

\subsection{Saturated Presentations}

\begin{definition}[\(L\)-saturated presentation]
\label{def:rule_saturated_presentation}
For a length bound \(L\), a presentation \(D\) is \emph{\(L\)-saturated} for
a target grammar \(G^*\) if
\[
D\cap\Sigma^{\le L}=L(G^*)\cap\Sigma^{\le L}.
\]
When the bound is fixed or clear from context, we simply say that \(D\) is
\emph{saturated}.
\end{definition}

\begin{proposition}[Fitness preservation for target subgrammars]
\label{prop:fitnessPreservation}
Let \(D\) be \(L\)-saturated for \(G^*\), and let \(G'\subseteq G^*\). If
\(G'\) generates at least one string within \(L\), then \(f_L(D\mid G')=1\).
\end{proposition}

\begin{proof}
Since \(G'\subseteq G^*\), we have \(L(G')\subseteq L(G^*)\). Hence
\[
L(G')\cap\Sigma^{\le L}
\subseteq
L(G^*)\cap\Sigma^{\le L}.
\]
By \(L\)-saturation,
\[
L(G^*)\cap\Sigma^{\le L}=D\cap\Sigma^{\le L}.
\]
Therefore every string generated by \(G'\) within the horizon is present in
\(D\). Thus \(EC_i(D\mid G')=GC_i(G')\) for all \(i\le L\). Since \(G'\)
generates at least one string within \(L\), the denominator of
\(f_L(D\mid G')\) is nonzero, and so \(f_L(D\mid G')=1\).
\end{proof}

In particular, the target grammar itself has finite-horizon fitness \(1\), and
so does any subgrammar of the target that generates at least one string within
the horizon. This idealized property is the formal basis for the search plateau: under a saturated presentation, every subgrammar that generates at least
one string within the horizon remains maximally fit over the fixed finite
universe. Conversely, any rule extension that generates a string in
\(\Sigma^{\le L}\setminus D\) fails the fitness test at that horizon, and by
Proposition~\ref{prop:overgeneration_persistence} later rule additions cannot
remove that overgeneration.

\begin{proposition}[Coverage bound for witnessed chains]
\label{prop:witnessed_chain_coverage}
Let
\[
G_0\subset G_1\subset\cdots\subset G_m
\]
be a monotone chain of grammars, and let \(L\) be a finite length bound.
Suppose the chain is \(L\)-witnessed: every rule of \(G_0\) has a witness in
\(D\) of length at most \(L\), and for each \(i<m\), every rule newly added in
\(G_{i+1}\setminus G_i\) occurs in a derivation, under \(G_{i+1}\), of some
string \(u_i\in D\) with \(|u_i|\le L\). Then
\[
\ell^*(G_i)\le L
\qquad\text{for all }i.
\]
\end{proposition}

\begin{proof}
The claim holds for \(G_0\) by assumption. Assume \(\ell^*(G_i)\le L\). Every
old rule \(r\in G_i\) keeps its previous witness in \(G_{i+1}\), since rule
addition is monotone. Hence its shortest witness length in \(G_{i+1}\) is
still at most \(L\). Every new rule in \(G_{i+1}\setminus G_i\) is witnessed,
under \(G_{i+1}\), by some string \(u_i\in D\) with \(|u_i|\le L\). Therefore
every rule of \(G_{i+1}\) has a witness of length at most \(L\), so
\(\ell^*(G_{i+1})\le L\). Induction gives the result.
\end{proof}

\subsection{Commutativity of Shortest-String Discovery}
\label{sec:commutativity}

The search space is the finite lattice of rule sets ordered by rule
addition. In a greedy traversal of this lattice, a central risk is that
committing to one rule increment might prevent the learner from reaching a
target-preserving path. The next result shows that, under saturation, shortest
unparsed evidence strings can be discovered in either order along such a path.

Let \(G_R\subseteq G^*\) be a fit intermediate hypothesis. The evidence strings
not yet parsed by \(G_R\) are
\[
U_R(D)=D\setminus L(G_R).
\]
A string \(u\in U_R(D)\) is a \emph{shortest unparsed evidence string} if
\[
|u|=\min\{|v|:v\in U_R(D)\}.
\]

Before stating the theorem, we fix the local terminology. A rule increment
\(\Delta\) is \emph{target-preserving} at \(G_R\) if
\[
\Delta\subseteq G^*\setminus G_R
\]
and \(G_R\cup\Delta\) remains fit. Such an increment \emph{explains} an
unparsed evidence string \(u\in U_R(D)\) if
\[
u\in L(G_R\cup\Delta).
\]
The \emph{incrementality} of a target-preserving chain is the largest size of
an increment needed along the chain. The search radius \(k\) is the maximum
increment size considered by the learner from a fit root.

\begin{theorem}[Commutativity of shortest-string discovery]
\label{thm:commutativity_discovery}
Assume \(D\) is \(L\)-saturated for \(G^*\), and let \(G_R\subseteq G^*\) be a
fit intermediate hypothesis. Let \(u,v\in U_R(D)\) be shortest unparsed
evidence strings explained by target-preserving increments
\(\Delta_u,\Delta_v\subseteq G^*\setminus G_R\). Then discovering
\(\Delta_u\) before \(\Delta_v\), or \(\Delta_v\) before \(\Delta_u\),
preserves a fit path to the same union
\[
G_R\cup\Delta_u\cup\Delta_v.
\]
\end{theorem}

\begin{proof}
Both \(G_R\cup\Delta_u\) and \(G_R\cup\Delta_v\) are subgrammars of \(G^*\).
Since \(\Delta_u,\Delta_v\subseteq G^*\setminus G_R\), the union
\(G_R\cup\Delta_u\cup\Delta_v\) is also a subgrammar of \(G^*\). By
Proposition~\ref{prop:fitnessPreservation}, all three grammars remain maximally
fit within the fixed saturated horizon, provided they generate at least one
string within \(L\).

Suppose \(\Delta_u\) is discovered first. Rule addition is monotone, so
\(G_R\cup\Delta_u\) can only parse additional evidence relative to \(G_R\). If
\(v\) is parsed already, then the evidence it supplied has been covered. If
\(v\) remains unparsed, adding \(\Delta_v\) later yields
\(G_R\cup\Delta_u\cup\Delta_v\), a superset of \(G_R\cup\Delta_v\); hence the
derivation of \(v\) available in \(G_R\cup\Delta_v\) remains available. The
same argument holds with \(u\) and \(v\) exchanged.
\end{proof}

\subsection{Search Implications}

The search consequence is that target-preserving paths remain viable under the
conditions above. Unproductive, unreachable, or redundant rule additions may
leave the language unchanged; these are the flat regions later controlled by
the incrementality parameter \(k\). The monotonicity and commutativity results show why a
greedy learner can search locally around fit hypotheses: target-preserving
rule additions remain on the fit plateau under saturation, while
rule additions that introduce unsupported short strings leave that plateau.

The learning algorithm in the next section makes this strategy concrete. It
searches over bounded rule increments from fit roots, uses shortest unparsed
evidence strings to guide candidate generation, and separately checks coverage
of the full observed presentation.

\section{The Learning Algorithm}
\label{sec:Learning Algorithm}

This section describes the hypothesis representation, the greedy search loop,
and the filters used to enumerate and evaluate candidate grammars.

\subsection{Grammar Interface and Hypothesis Space}

The learner operates on a finite presentation \(D\subseteq P^*\) of
preterminal strings, where \(P\) is the set of observed preterminal categories.
The word-to-preterminal lexicon is external to the learner; the hypothesis
space below concerns only the structural grammar over preterminals. Let \(N\)
be the finite set of structural nonterminals. The finite structural universe is
\[
\mathcal{R}_{\mathrm{core}}
=\{S\to A:A\in N\}\cup\{A\to BC:A,B,C\in N\},
\]
and the structural-preterminal interface is
\[
\mathcal{R}_{\mathrm{pre}}=\{A\to p:A\in N,\ p\in P\}.
\]

Thus the searched normal form contains distinguished start projections, binary
structural rules, and preterminal interface rules, but no arbitrary structural
unit rules. This is a Chomsky-normal-form-style restriction: apart from the separate start
and \(\epsilon\) cases, CFGs can be converted to binary form without changing
their string language \citep{hopcroft1979introduction}.
A run fixes maximum values for the number of structural nonterminals and
rules. If \(\epsilon\) belongs to
the target language, it is handled as a separate start case outside the
structural rule search.

For a core rule set \(C\subseteq\mathcal{R}_{\mathrm{core}}\), the learner
maintains a partial functional mapping
\[
\sigma_C:\operatorname{dom}(\sigma_C)\subseteq N\to P.
\]
The mapping is not required to be injective: distinct structural nonterminals
may map to the same preterminal. A hypothesis is \(H=(C,\sigma_C)\). Its
coverage and fitness are computed over the induced
preterminal-string language \(L_P(H)\subseteq P^*\), using the grammar-count
and evidence-count vectors defined in Section~\ref{sec:Fitness Function}.

This separation is algorithmic, not only notational. The search enumerates core
rule sets without committing in advance to all structural-preterminal
assignments. When a candidate is tested against the next shortest unparsed
string, CKY preterminal abduction infers only the functional interface
extensions needed for a derivation of that string. Thus the learner avoids the
full cross-product of core candidates with complete mappings \(N\to P\), and
lets the next unparsed string impose bottom-up constraints on the
structural-preterminal interface.

\subsection{Search Overview}

Algorithm~\ref{alg:main_search} summarizes the learner. The learner first
performs a bootstrap search from the empty rule set, up to an initialization
radius \(k_0\), to find incomplete fit hypotheses that parse at least one
evidence string. These hypotheses initialize the root queue \(Q\).

The recognition and counting tests depend only on the set \(D\). The search
schedule derives an order \(\hat{D}\) from the observed string types:
preterminal strings are sorted by nondecreasing length, with ties preserving
their relative order in the original presentation. At a root \(R\), the next
unparsed string is the first string in \(\hat{D}\) not recognized by \(R\).

For compactness, the pseudocode uses the batched implementation view. A search
node is written \(B=(C,\mathcal{M}_C)\), where \(C\) is a core rule set and
\(\mathcal{M}_C\) is the current list of compatible partial mappings. Each
\(\sigma\in\mathcal{M}_C\) represents the mapped hypothesis \((C,\sigma)\).
Parsed-evidence vectors and next-unparsed strings are maintained per mapping,
while the canonical adjacency search over \(C\) is shared.

From each root, the learner performs an inner breadth-first search over rule
additions of size at most \(k\). The inner search is guided by the relevant
next-unparsed evidence strings: candidates are generated canonically, wildcard
guidance focuses the next adjacency step, compatible structural-preterminal
mappings are abduced, and the resulting hypotheses are evaluated.

\begin{algorithm}[t]
\caption{Greedy Grammar Induction}
\label{alg:main_search}
\begin{algorithmic}[1]
\Require preterminal presentation \(D\), initialization radius \(k_0\), search
radius \(k\), post-solution depth slack \(\rho\), Pareto-ribbon width \(\eta\),
rule universe \(\mathcal{R}_{\mathrm{core}}\)
\State derive the sorted presentation order \(\hat{D}\) from \(D\)
\State initialize solution tracker \(\mathcal{T}\) and root queue \(Q\)
\State initialize \(Q\) with incomplete fit roots found by bootstrap search from
the empty rule set up to radius \(k_0\)
\While{\(Q\) is not empty}
    \State dequeue a fit root \(R=(C_R,\mathcal{M}_R)\)
    \label{alg:dequeue_root}
    \If{\(\mathcal{T}\) contains a complete solution of minimum depth
    \(d_{\min}\) and \(\operatorname{depth}(R)>d_{\min}+\rho\)}
        \State continue
    \EndIf
    \State for each active \(\sigma\in\mathcal{M}_R\), select the first
    \(u_{R,\sigma}\in\hat{D}\) not parsed by \((C_R,\sigma)\)
    \label{alg:select_root_string}
    \State initialize inner frontier \(F_0\gets\{R\}\)
    \label{alg:init_inner_frontier}
    \For{\(d=0,\ldots,k\)}
        \State initialize next frontier \(F_{d+1}\gets\emptyset\)
        \For{each batched node \(B=(C,\mathcal{M}_C)\in F_d\)}
            \State abduce all functional mapping extensions that parse the
            mapping-specific next-unparsed strings
            \label{alg:preterminal_abduction}
            \For{each resulting mapped hypothesis \(H=(C,\sigma)\)}
                \State compute \(\ell^*(H)\), \(\mathbf{GC}_{\ell^*(H)}(H)\),
                \(\mathbf{EC}_{\ell^*(H)}(D\mid H)\), and \(f(D\mid H)\)
                \label{alg:evaluate_shape}
                \If{\(H\) contains an unproductive nonterminal or
                \(f(D\mid H)<1\)}
                    \State reject \(H\)
                \ElsIf{\(H\) passes tracker filtering}
                    \If{\(H\) recognizes all of \(D\)}
                        \State record \(H\) as a complete solution in
                        \(\mathcal{T}\), subject to the Pareto-ribbon policy
                        \label{alg:insert_global}
                    \Else
                        \State enqueue \(H\) into \(Q\) as a future fit root
                        \label{alg:enqueue_root}
                    \EndIf
                \EndIf
            \EndFor
            \If{\(d<k\)}
                \State generate the canonical adjacents of \(C\)
                \label{alg:canonical_adjacents}
                \State retain adjacents satisfying the wildcard guide for at
                least one active mapping
                \label{alg:wildcard_guidance}
                \State use all canonical adjacents if guidance fails
                \label{alg:guided_fallback}
                \State add the surviving adjacents, with their surviving
                mapping lists, to \(F_{d+1}\)
            \EndIf
        \EndFor
    \EndFor
\EndWhile
\State \Return complete solutions stored in \(\mathcal{T}\)
\end{algorithmic}
\end{algorithm}

Algorithm~\ref{alg:main_search} is written in sequential form. In the
implementation, candidate processing within a fixed inner frontier \(F_d\) is
parallelized across independent nodes; this reduces wall-clock time but does
not change the generated candidate set or the correctness argument.

Only incomplete fit hypotheses are enqueued as roots, so every active mapping at
a dequeued root has a next unparsed string. After the first complete solutions
are found, the post-solution slack \(\rho\) limits further expansion: roots
deeper than \(d_{\min}+\rho\), where \(d_{\min}\) is the minimum depth of a
complete solution, are not expanded. The Pareto-ribbon width \(\eta\) controls
retention among complete fit hypotheses, as described in
Section~\ref{sec:evaluation_parsimony}.

The remaining subsections unpack the main operations in
Algorithm~\ref{alg:main_search}. Candidate evaluation computes the
rule-coverage bound, count vectors, and fitness score. Hypotheses are rejected
if they contain an unproductive nonterminal or generate unsupported strings up
to their rule-coverage bound. Complete fit hypotheses are recorded as global
solutions, while incomplete fit hypotheses become future roots. Canonical
adjacency and wildcard guidance generate the next inner frontier, and CKY
preterminal abduction supplies compatible interface mappings for candidates in
that frontier.
\subsection{Canonical Adjacency}

Canonical adjacency implements line~\ref{alg:canonical_adjacents} on the core
rule set \(C\) of a batched node. Its purpose is to enumerate core rule sets
without duplicate paths caused by different orders of adding the same rules.
Fix a total order
\[
\mathcal{R}_{\mathrm{core}}=\{r_1,\ldots,r_n\}.
\]
Each core-rule node stores the largest index \(m\) used on its canonical path
from the current root, and may add only rules \(r_j\) with \(j>m\). Thus a rule
set \(\{r_{j_1},\ldots,r_{j_t}\}\) is generated only in increasing index
order, rather than once for every permutation of its rules.

The ordering is paired with a canonical naming convention for nonterminals.
Names \(X_1,X_2,\ldots\) are introduced without gaps, and a non-start
nonterminal may serve as the left-hand side of a binary rule only after it has
already been introduced by a start projection or as a right-hand-side symbol.
These gates exclude orphaned rules and strongly equivalent grammars that differ
only by gaps or skipped names. They do not remove any language-level solution:
any reachable productive grammar has an equivalent representative obtained by
discarding unreachable rules and renaming nonterminals in order of first
introduction.

Canonical adjacency therefore removes permutation redundancy while preserving
the admissible generated hypothesis space, modulo harmless renaming of
nonterminals. The formal exactness statement for canonical adjacency is given
in Appendix~\ref{app:canonical_guided}.

\subsection{Guided Generation and Preterminal Abduction}

Canonical adjacency controls duplicate enumeration, but most nearby rule
extensions cannot help parse the next shortest unparsed evidence string for any
active root mapping. The guided-generation step therefore focuses the inner
frontier on rules that can participate in a derivation of one of those selected
strings.

Fix a fit root \(R\), a current batched node \(B=(C,\mathcal{M}_C)\) with
\(C=R\cup\Delta\), an active mapping \(\sigma\in\mathcal{M}_C\), its selected
string \(u_{R,\sigma}\), and a canonical adjacent rule \(c_t\). The wildcard
guide asks whether \(u_{R,\sigma}\) has a wildcard-relaxed derivation using
\(c_t\) and the current delta \(\Delta\). The wildcard relaxation allows
currently unproductive non-start nonterminals to stand for one nonempty span of
the selected string. This tests whether the adjacent rule can help explain the
next evidence string before all downstream structure has been fixed.

The guide is an exact filter for its stated survivor predicate when its dynamic
program applies. Candidates rejected by a successful guide cannot be the next
increment explaining the selected string under the corresponding mapped root.
If the guide cannot be applied safely, the learner falls back to the full
canonical adjacent list. Thus guided generation is a search filter, not a
fitness test. Appendix~\ref{app:canonical_guided} defines the survivor predicate
and proves the exactness of successful guided generation.

The survivor predicate is prefix-local. A candidate is retained only when it can
participate in a wildcard derivation of a selected string together with the
rules already present in the current root-local delta \(\Delta\). The guide does
not assume that rules are added in parse-tree order: a retained candidate may
appear above, below, or beside earlier delta rules in the eventual derivation.
Completeness for guided search is therefore conditional on the needed canonical
prefixes surviving this predicate, except when the guide falls back to the full
canonical list.

Nodes that survive guidance are passed to CKY preterminal abduction
\citep{kasami1966efficient,younger1967recognition}. For a selected string
\(u_{R,\sigma}\) and current partial mapping \(\sigma\), the CKY-based
abduction step enumerates all functional mapping extensions \(\tau\) for which
\(\sigma\cup\tau\) permits a derivation of that string. This step infers the
structural-preterminal interface only; the preterminal-to-word lexicon remains
external.

The two filters play complementary roles. Wildcard guidance asks which
adjacents can participate in a derivation of a selected string for at least one
active mapping. Preterminal abduction asks which interface rules make such a
derivation compatible with the observed preterminals. Together they avoid
enumerating the full cross-product of rule extensions and complete
structural-preterminal mappings.

\subsection{Evaluation, Parsimony, and Strong-Equivalence Filtering}
\label{sec:evaluation_parsimony}

Candidate evaluation first checks productivity: every nonterminal must derive
at least one preterminal string. Productive candidates are then evaluated by
the fitness test at their rule-coverage bound. This rejects hypotheses that
generate unsupported preterminal strings within their comparison universe. The
candidate is also checked for recognition of the entire observed presentation,
including strings longer than \(\ell^*(H)\). Hypotheses that pass the fitness
test but do not yet cover all of \(D\) become future roots; hypotheses that also
recognize all of \(D\) are complete fit hypotheses.

Among complete fit hypotheses, the learner uses a two-part MDL-style selection
logic \citep{rissanen1978modeling,grunwald2007minimum}. For a hypothesis
\(H=(C,\sigma_C)\), the rule-set size
\[
d(H)=|C|
\]
serves as a proxy for model description length. The grammar growth rate
\(\lambda(H)\), computed as the reciprocal of the estimated radius of
convergence of the grammar's generating-function system
\citep{flajolet1987analytic}, serves as an asymptotic narrowness measure and
as a proxy for the data-given-model term. Related algorithmic results for
context-free language growth are given by
\citet{gawrychowski2010growth}. Lower growth means that the grammar leaves
fewer alternative preterminal strings available and is therefore more specific.
This is distinct from the fitness score \(f(D\mid H)\), which is a feasibility
constraint: a hypothesis is fit only if this score equals \(1\).

For ambiguous grammars, \(\lambda(H)\) is the growth rate of the
derivation-counting generating system. It upper-bounds, and may strictly
exceed, the growth rate of the weak language, because the same string may have
multiple derivations. We therefore use \(\lambda(H)\) only as a parsimony and narrowness proxy among
complete fit hypotheses; it is not a test of weak equivalence and not a
likelihood score.

Rather than flattening \(d(H)\) and \(\lambda(H)\) into a single scalar MDL
score, the learner maintains a Pareto-ribbon archive of complete fit
hypotheses. The ribbon width \(\eta\ge 0\) controls pruning. A candidate
\((d,\lambda)\) is pruned only if some accepted complete hypothesis
\((d',\lambda')\) satisfies
\[
(1+\eta)d'\le d
\qquad\text{and}\qquad
(1+\eta)\lambda'\le \lambda .
\]
When \(\eta=0\), this is ordinary Pareto dominance. When \(\eta>0\), the test
is conservative and may retain complete hypotheses that are slightly worse than
the current frontier. The retained archive therefore represents a ribbon of
size--growth tradeoffs rather than a single imposed optimum.

The algorithm nevertheless leans toward parsimony. Let \(d_0\) be the minimum
rule-set size of any complete fit hypothesis found during the run. Once such
solutions exist, the post-solution slack \(\rho\) limits how far the search
continues beyond this shortest complete fit depth: roots deeper than
\(d_0+\rho\) are not expanded. Smaller values of \(\rho\) favor early stopping
near the most compact complete hypotheses; larger values allow deeper Pareto
competitors to be discovered. The ribbon width \(\eta\) and post-solution slack \(\rho\) are model-selection
parameters: they control how much of the size--growth frontier is explored and
retained, after hypotheses have passed the fitness and recognition tests.

The tracker is used for all fit hypotheses, not only complete ones. It filters
hypotheses under strong equivalence before incomplete hypotheses become future
roots and before complete fit hypotheses are recorded as global solutions.
Strong equivalence means equality of the full mapped hypotheses, including both
core rules and preterminal-interface rules, up to renaming of nonterminals, with
preterminal labels fixed. A
color-refinement signature is used as a cheap first pass, and exact nauty
canonicalization \citep{mckay2014practical} is invoked only on signature
collisions.

\section{Computational Complexity}
\label{sec:Complexity}

The learner has two main complexity sources: bounded enumeration of the rule
lattice, and candidate evaluation whose cost depends on the number of mapping
extensions and generated strings actually produced. Let \(m=|N|\) be the number
of nonterminals, let
\[
R_c=|\mathcal{R}_{\mathrm{core}}|=O(m^3),
\]
let \(k\) be the search radius, let \(L=\ell^*(G)\) be the rule-coverage bound
of a candidate grammar, let \(p=|P|\) be the number of preterminals, and let \(n\) be the maximum length of an
evidence string.

For a fixed root, unguided canonical enumeration of increments of size at most
\(k\) generates at most
\[
\sum_{i=1}^{\min(k,R_c)}\binom{R_c}{i}.
\]
For each fixed \(k\), this is \(O(R_c^k)=O(m^{3k})\). Thus the enumeration
component is XP in \(k\): it is polynomial on every fixed-\(k\) slice, but the
polynomial degree depends on \(k\)
\citep{downey2013fundamentals}. Canonical adjacency removes permutation
redundancy by generating each rule set once; it does not change this worst-case
subset bound.

Wildcard-guided generation applies a batched chart computation to selected
next-unparsed evidence strings. For current delta size \(k_\Delta\le k\),
selected-string length \(B\le n\), and \(q\) candidate rules, one guide pass for
a fixed structural-preterminal mapping stores
\[
O\!\left(m\,2^{k_\Delta}(B+1)^2\left\lceil q/64\right\rceil\right)
\]
machine words. The factor \((B+1)^2\) is the number of chart intervals over
string positions \(0,\ldots,B\). The exponential factor \(2^{k_\Delta}\) comes
from recording which root-local delta rules have been used in the wildcard
derivation. The bitset factor processes candidate rules word-parallel; it
reduces constants but does not change the worst-case exponential dependence on
the local delta size.

For each surviving candidate, CKY preterminal abduction has the usual cubic
span dependence in the selected-string length, together with a term proportional
to the number of functional mapping extensions returned. This dependence is
output-sensitive: the procedure explicitly enumerates the compatible mapping
extensions, so its running time must scale with how many such extensions exist.

Candidate evaluation then materializes the distinct preterminal strings
generated up to the rule-coverage bound \(L\). If
\[
M_L(G)=|L_P(G)\cap P^{\le L}|,
\]
then the exact count computation is output-sensitive in \(M_L(G)\), with the
crude worst-case bound
\[
M_L(G)\le \sum_{i=0}^{L}p^i=O(p^L)
\]
for \(p>1\). Thus this step is tractable only when the rule-coverage bound
remains sufficiently small and the bounded generated language can be
materialized.

Full evidence recognition is a standard CKY pass over the finite presentation,
bounded coarsely by \(O(|D|\,n^3\,|G|)\) after indexing. Growth-rate estimation,
Pareto-ribbon checks, and strong-equivalence tracking are applied to hypotheses
that survive productivity and fitness filtering; full-presentation recognition
determines whether such a hypothesis is recorded as complete or kept as a
future root. If \(\mathcal{O}\) denotes the fit roots explored during a run and
\(T_{\mathrm{eval}}\) denotes the staged candidate-evaluation cost, a coarse
run-dependent bound is
\[
O(|\mathcal{O}|\,R_c^k\,T_{\mathrm{eval}}),
\]
where \(T_{\mathrm{eval}}\) absorbs guided generation, abduction, recognition,
materialization of generated strings up to \(L\), growth-rate estimation, and
tracker operations, including the output-sensitive terms.

\section{Verification and Conditional Recovery}
\label{sec:Learnability}

The learner is not a distribution-free identification procedure for arbitrary
context-free grammars. Its guarantees concern a finite hypothesis space, a
finite presentation, and the bounded overgeneration check defined by fitness.
This section separates what accepted hypotheses have been verified to satisfy
from the additional reachability assumptions under which weak recovery follows.

\subsection{Verification of Accepted Hypotheses}

A mapped hypothesis \(H=(C,\sigma_C)\) is \emph{fit} for \(D\) if
\[
f(D\mid H)=1.
\]
Thus \(H\) generates no unsupported preterminal string up to its own
rule-coverage bound \(\ell^*(H)\).

A reported complete hypothesis is a fit hypothesis that also recognizes the
entire presentation. Hence every reported complete hypothesis satisfies
\[
D\subseteq L_P(H)
\]
and
\[
L_P(H)\cap P^{\le \ell^*(H)}
\subseteq
D\cap P^{\le \ell^*(H)}.
\]
These are finite-presentation verification conditions. They do not imply weak
equivalence to an unknown target, nor do they give a distribution-free
generalization guarantee beyond the rule-coverage bound. Conditional weak
recovery requires the reachability assumptions stated next.

\subsection{Conditional Reachability}

Let \(G^*\) be a target grammar over preterminal strings, and let \(L\) be a
finite saturation horizon. We say that \(G^*\) is \emph{\((L,k)\)-reachable}
for a presentation \(D\), under the fixed symbol and rule bounds, if \(D\) is
\(L\)-saturated for \(G^*\) and there exists a monotone chain of productive
mapped hypotheses
\[
H_0\subset H_1\subset\cdots\subset H_m
\]
in the hypothesis space such that \(H_m\) is weakly equivalent to \(G^*\). Here
\(H_i=(C_i,\sigma_i)\), and inclusion means extension of both the core rule set
and the partial structural-preterminal mapping.

The chain is required to be visible to the local search. First, \(H_0\) is
reachable by the bootstrap search, parses at least one evidence string, and has
all of its rules witnessed by strings in \(D\) of length at most \(L\). Second,
for each \(i<m\), \(H_i\) is incomplete on \(D\), and \(H_{i+1}\) extends
\(H_i\) by at most \(k\) core rules, together with a functional extension of
the mapping, so as to parse the next unparsed evidence string selected at
\(H_i\). Every newly added rule must occur in a derivation of that selected
string under \(H_{i+1}\). Third, the increment must be
wildcard-guide-admissible: when its core rules are added in canonical order
from \(H_i\), each required adjacent satisfies the wildcard survivor predicate
at its prefix, unless wildcard-guided generation falls back to the full
canonical list. When several shortest unparsed strings are available, the fixed
presentation order chooses one witness; same-length target-preserving
alternatives commute as in Theorem~\ref{thm:commutativity_discovery}.

This definition does not assume that the learner succeeds. It specifies a
bounded, evidence-witnessed chain to a target-equivalent hypothesis. By
Proposition~\ref{prop:witnessed_chain_coverage}, the witnessed-chain condition
implies
\[
\ell^*(H_i)\le L
\qquad\text{for all }i.
\]
Since \(H_i\subseteq H_m\) and \(H_m\) is weakly equivalent to \(G^*\), every
preterminal string generated by \(H_i\) is generated by \(G^*\). Because \(D\)
is \(L\)-saturated for \(G^*\), Proposition~\ref{prop:fitnessPreservation}
implies that every \(H_i\) is fit. Thus the chain lies on the fit plateau used
by the learner.

\subsection{Conditional Weak Recovery}

\begin{theorem}[Conditional weak recovery]
\label{thm:conditional_weak_recovery}
If a target grammar \(G^*\) is \((L,k)\)-reachable for \(D\) under the fixed
symbol and rule bounds, and if exact wildcard-guided generation and abduction
are used, then the idealized learner reaches a hypothesis weakly equivalent to
\(G^*\).
\end{theorem}

\begin{proof}
By \((L,k)\)-reachability, there is a monotone chain
\[
H_0\subset H_1\subset\cdots\subset H_m
\]
in the hypothesis space such that \(H_m\) is weakly equivalent to \(G^*\). The
bootstrap search reaches \(H_0\).

We prove by induction that the learner reaches every \(H_i\) on the chain. The
base case is \(H_0\). Suppose the learner has reached \(H_i\) with \(i<m\).
By the consequence above, \(H_i\) is fit. Since \(H_i\) is incomplete on \(D\),
it is eligible to serve as a root. At that root, the search selects the next
unparsed evidence string specified in the reachability condition.

The increment from \(H_i\) to \(H_{i+1}\) has at most \(k\) core rules.
Canonical adjacency enumerates this increment in canonical order within the
bounded neighborhood. Wildcard-guide admissibility ensures that each required
adjacent survives at its prefix, unless wildcard-guided generation falls back
to the full canonical list. CKY preterminal abduction enumerates the functional
mapping extension needed for the selected string. Hence the candidate
\(H_{i+1}\) is generated. The consequence above gives that \(H_{i+1}\) is fit,
so it is accepted by the fitness test. If \(i+1<m\), it is incomplete and
becomes a future root; if \(i+1=m\), it recognizes the full presentation and is recorded as a complete
fit hypothesis. Thus the
learner reaches \(H_{i+1}\). By induction, it reaches \(H_m\), which is weakly
equivalent to \(G^*\).
\end{proof}

The theorem is a recovery result for a finite search problem, not an
identification-in-the-limit theorem. A saturated presentation supplies the bounded witnesses needed to keep
target-preserving paths visible to the learner; the substantive reachability
parameter is the increment size \(k\), which controls how locally such a path
must decompose. The exact search mechanisms then preserve any such witnessed
bounded path.

In practice, the user chooses how far to continue the search and how much of the
size--growth frontier to retain, subject to computational and time budgets. The
post-solution slack \(\rho\) and Pareto-ribbon width \(\eta\) implement these
choices. They do not change the reachability condition, but recovery in a
particular run requires that they not prune the reachable chain before it
reaches \(H_m\).

\section{\texorpdfstring{Sample Requirements for \(L\)-Saturation}{Sample Requirements for L-Saturation}}
\label{sec:SampleComplexity}

The reachability guarantees in Section~\ref{sec:Learnability} assume an
\(L\)-saturated presentation. This is a finite-coverage assumption. In
random samples, saturation is not automatic: every valid target string up to
length \(L\) must be observed at least once. This section makes the associated
sample requirement explicit.

Let
\[
S_L=L(G^*)\cap\Sigma^{\le L},
\qquad
M_L=|S_L|.
\]
For a distribution \(\mathcal{D}\) over \(L(G^*)\), define
\[
p_{\min,L}=\min_{w\in S_L}\Pr_{\mathcal{D}}(w),
\]
assuming \(p_{\min,L}>0\).

\begin{theorem}[High-probability saturation bound]
\label{thm:sample_complexity}
Let \(D\) contain \(N\) independent samples from \(\mathcal{D}\). Then
\[
\Pr[D\ \text{is not \(L\)-saturated}]
\le
M_L e^{-Np_{\min,L}}.
\]
Consequently, it suffices to take
\[
N\ge \frac{1}{p_{\min,L}}\ln\left(\frac{M_L}{\delta}\right)
\]
to obtain \(L\)-saturation with probability at least \(1-\delta\).
\end{theorem}

\begin{proof}
A fixed string \(w\in S_L\) is absent from \(N\) independent samples with
probability \((1-\Pr_{\mathcal{D}}(w))^N\le e^{-N\Pr_{\mathcal{D}}(w)}\).
Taking a union bound over all \(M_L\) strings gives
\[
\Pr[\exists w\in S_L:w\notin D]
\le
\sum_{w\in S_L}e^{-N\Pr_{\mathcal{D}}(w)}
\le
M_L e^{-Np_{\min,L}}.
\]
Solving \(M_L e^{-Np_{\min,L}}\le\delta\) yields the stated sufficient
condition.
\end{proof}

As a stylized heavy-tailed model, suppose the distribution is Zipfian on
\(S_L\): after ranking strings by probability,
\[
\Pr(w_i)=\frac{i^{-s}}{\mathcal{H}_{M_L,s}},
\qquad
\mathcal{H}_{M_L,s}=\sum_{j=1}^{M_L}j^{-s}.
\]
Then
\[
p_{\min,L}=\frac{M_L^{-s}}{\mathcal{H}_{M_L,s}},
\]
and the sufficient condition becomes
\[
N\ge
M_L^s\mathcal{H}_{M_L,s}
\ln\left(\frac{M_L}{\delta}\right).
\]

For \(s=0\), this recovers the uniform coupon-collector bound
\[
N\ge M_L\ln\left(\frac{M_L}{\delta}\right).
\]
For \(s=1\),
\[
N\ge M_L\mathcal{H}_{M_L,1}
\ln\left(\frac{M_L}{\delta}\right)
=
O\!\left(M_L\log M_L\log\frac{M_L}{\delta}\right).
\]

Because \(M_L\le \sum_{i=0}^L|\Sigma|^i\), exact saturation can require a
sample size exponential in \(L\). Thus \(L\)-saturation is best understood as
a sufficient finite-coverage condition for the recovery theorem, not as a
distribution-free consequence of positive sampling. Whether naturalistic input
provides this kind of bounded coverage for the relevant grammar fragments is an
empirical question. The experiments in Section~\ref{sec:EmpiricalEvaluation}
therefore do not require exact \(L\)-saturation; instead, they use a weaker
missing-mass stopping rule.

\section{Empirical Evaluation}
\label{sec:EmpiricalEvaluation}

We evaluate the learner on a targeted suite of 31 runs. A run consists of a
target grammar, a fixed external lexicon, and a search setting. The suite tests
recursive structure, matching dependencies, ambiguity, and
natural-language-inspired adjunction. It includes English-like and
subject-object-verb fragments, adjectival recursion, PP adjunction, ambiguous
PP adjunction, Dyck-\(k\) languages for \(1\le k\le4\), odd, even, and general
palindrome languages, the non-substitutable language \(\{a^n b^n:n\ge1\}\),
and the inherently ambiguous union
\(\{a^n b^n c^m:n,m\ge1\}\cup\{a^m b^n c^n:n,m\ge1\}\).

All runs use positive evidence generated from the target grammar after applying
the fixed external lexicon. Sampling uses the implementation's power-law
distribution with a fixed random seed, so the presentation \(D\) is
reproducible. Sampling stops when a Good--Turing-style missing-mass estimate
falls below the stopping threshold
\citep{good1953population,good1956number}. As noted in Section~\ref{sec:SampleComplexity}, the experiments do not require
exact \(L\)-saturation. The empirical question is whether the resulting
presentations are witness-rich enough for recovery in the tested families; in all 31 runs, they were. 

Several English-style targets are examined under two different ways of encoding
verb subcategorization at the syntactic-lexical interface. In the
subcategorized versions, verbal preterminals distinguish intransitive,
transitive, PP-taking, and sentential-complement-taking verbs. In the
single-category versions, these distinctions are removed and all verbs share a
single preterminal \(V\). This second encoding tests recovery when argument
structure is not represented by separate verbal preterminals. Surface strings in
the logs should therefore be read as lexical realizations of formal
preterminal strings, with no selectional or semantic restrictions imposed
beyond the CFG target.

In all 31 runs, grammar-level analysis establishes weak equivalence between the
reported output grammar and the corresponding target language. The
implementation also performs a bounded sampled stress test through length 30,
drawing positive examples from the target grammar and negative examples
rejected by the target grammar; this stress test is diagnostic rather than a
proof of equivalence.

In Table~\ref{tab:test_suite_summary}, \(\ell^*(G)\) refers to the reported
target-equivalent output grammar in that run, not to all fit hypotheses explored
by the learner. Wall-clock times were measured for the parallel C\# implementation on Windows
10 Home, using an AMD Ryzen 7 3700X 8-core processor with 16 logical processors
and 64 GB RAM.

\begin{table}[t]
\centering
\small
\begin{tabular}{lrccr}
\hline
Benchmark family & Runs & \(k\) & \(\ell^*\) & Time \\
\hline
English fragments, subcategorized verbs
    & 8 & 1--2 & 4 & \(0\mathrm{s}\)--\(9\mathrm{s}\) \\
English fragments, single verbal preterminal
    & 8 & 1--2 & 4 & \(1\mathrm{s}\)--\(15\mathrm{s}\) \\
SOV fragments
    & 4 & 1--2 & 4 & \(0\mathrm{s}\)--\(6\mathrm{s}\) \\
\shortstack[l]{Ambiguous PP-adjunction fragments,\\ subcategorized and single variants}
    & 2 & 2 & 4 & \(1{:}16\)--\(1{:}22\) \\
Dyck-\(k\), \(1\le k\le4\)
    & 4 & 2 & 4 & \(0\mathrm{s}\)--\(11\mathrm{s}\) \\
Palindrome languages
    & 3 & 1--2 & 3--4 & \(0\mathrm{s}\)--\(21\mathrm{s}\) \\
Non-substitutable \(a^n b^n\)
    & 1 & 1 & 4 & \(0\mathrm{s}\) \\
Inherently ambiguous \(a,b,c\) union
    & 1 & 3 & 5 & \(2{:}28{:}16\) \\
\hline
Total
    & 31 & 1--3 & 3--5 & \(2{:}33{:}08\) \\
\hline
\end{tabular}
\caption{Benchmark-suite summary. The table reports ranges over runs in each
family. The \(\ell^*\) column reports the rule-coverage bound of the
target-equivalent grammar listed for the run. Times are wall-clock runtimes,
rounded to seconds.}
\label{tab:test_suite_summary}
\end{table}

Appendix~\ref{app:recovered_grammars} lists representative recovered grammars
from the suite, including generated string counts inside the rule-coverage
horizon. The supplementary material contains the complete fixed-seed
test-suite logs and the C\# implementation of the learner.

The suite includes targets outside substitutability-based learnability
conditions. The even-palindrome benchmark, \(a^n b^n\), and the inherently
ambiguous union language test whether the bounded search-and-rejection
criterion can recover correct weak languages in cases where other
positive-identification results do not directly apply.

The suite therefore tests a different regime from standard positive-data
guarantees for CFG induction, which typically rely on restricted language
classes or enriched structural examples
\citep{clark2007polynomial,sakakibara1992efficient}. The present experiments
instead use positive unstructured strings and test bounded weak recovery under
the reachability and saturation assumptions of Section~\ref{sec:Learnability}.

The runtime pattern is consistent with the slice-wise complexity analysis in
Section~\ref{sec:Complexity}. Most runs finish quickly at \(k\le2\), while the
inherently ambiguous union example, the only run with \(k=3\), is substantially
more expensive. This reflects the expected combinatorial effect of increasing
the rule-increment radius, together with the larger number of compatible local
explanations in the ambiguous target.

\section{Limitations and Future Work}
\label{sec:future}

The current learner assumes an external word-to-preterminal lexicon and learns
the grammar over preterminal strings. This keeps the search problem finite and
explicit, but limits end-to-end applicability. Joint lexical and grammatical
induction is an important next step, and would add ambiguity both in sampling
and in parsing.

The fitness test is strict: a candidate that generates an unsupported short
preterminal string is rejected. Future work should replace or augment this test
with statistically calibrated slack for incomplete or noisy samples, especially
under heavy-tailed input. This connects to the sampling assumptions in
Section~\ref{sec:SampleComplexity}: the experiments use a Good--Turing-style
missing-mass stopping rule rather than requiring exact \(L\)-saturation
\citep{good1953population,good1956number,orlitsky2016optimal}. Whether
naturalistic input provides sufficient bounded coverage for the relevant
grammar fragments remains an empirical question.

Finally, the search parameters are currently chosen manually. The bootstrap
radius \(k_0\), incrementality radius \(k\), post-solution slack \(\rho\), and
Pareto-ribbon width \(\eta\) determine how much of the frontier is explored and
retained. A natural extension is an iterative-deepening version that increases
these parameters progressively, using held-out evidence, missing-mass estimates,
or frontier statistics for model selection.

\section{Conclusion}
\label{sec:Conclusion}

This paper studied positive-only context-free grammar induction over
preterminal strings. The main device is the rule-coverage bound
\(\ell^*(G)\), which gives each candidate grammar its own bounded comparison
universe. Within this universe, positive evidence can act as indirect negative
evidence: a grammar must parse the observed presentation, and it must not
generate unsupported preterminal strings up to its rule-coverage bound.

The resulting fitness function does more than score hypotheses. Under a
saturated presentation, it turns the rule lattice into a greedy search
landscape: target-preserving subgrammars stay on the fit plateau, while
extensions that generate unsupported short strings leave that plateau
permanently. This is what makes local rule addition principled rather than
merely heuristic.

The learner exploits this landscape by searching in bounded rule increments.
Canonical adjacency removes permutation redundancy, wildcard guidance focuses
local extensions on the next shortest unparsed evidence string, CKY
preterminal abduction infers the interface to observed preterminals, and
complete fit hypotheses are compared by size, grammar growth, Pareto-ribbon
filtering, and strong equivalence.

Across 31 benchmark runs, including English-style fragments, Dyck languages,
palindromes, \(a^n b^n\), and an inherently ambiguous union language, the
returned grammars are weakly equivalent to their targets by grammar-level
analysis, with enumeration through length 30 used as an additional bounded
check.

The contribution is therefore to turn positive-only CFG induction into a
finite recovery problem with a greedy search landscape: when the evidence
exposes a bounded path through the grammar lattice, the learner is guaranteed
to find a grammar weakly equivalent to the target.

\acks{The author received no external funding for this work and declares no competing interests.}

\newpage
\appendix

\section{Representative Recovered Grammars}
\label{app:recovered_grammars}

This appendix records a selection of representative grammars
returned by the benchmark suite. The full logs, including global optima, verification outputs, and
runtime traces, are provided in the supplementary material. The listings below
are recovered grammars whose weak equivalence to the corresponding target
grammar was checked by the grammar-level analysis reported in
Section~\ref{sec:EmpiricalEvaluation}.

The printed grammars use the implementation's compact display form. Internally
the learner remains in the lattice normal form defined in
Section~\ref{sec:Learning Algorithm}, with structural rules \(S\to X_i\),
\(X_i\to X_jX_k\), and structural-preterminal interface rules \(X_i\to p\).
For readability, the printer inlines this interface: if a variable
is mapped to a preterminal, the listing may show the preterminal directly
inside a rule. These displays are concise readable projections of the same
normal-form hypotheses, not a different hypothesis space.

The printed grammar-count vector lists
\(\mathbf{GC}_{\ell^*(G)}(G)\), namely the number of generated string types at
each length up to the rule-coverage bound. The printed ``total strings'' value
is
\[
M_{\ell^*(G)}(G)=|L_P(G)\cap P^{\le \ell^*(G)}|,
\]
the total number of generated preterminal string types inside that horizon; it
is not the size of the input presentation \(D\). The scalar rule-coverage bound
is printed separately. The empty string, when present, is handled by the
separate start case \verb|START -> \epsilon| and is not counted as a
rule-coverage witness.

\subsection{English-Style and SOV Examples}

\subsubsection*{English fragment (\(K=2\), global \#3)}
\begin{verbatim}
3. Global Optimum (Depth 7) Growth Rate: 1.594 [Indices: 39]:
\end{verbatim}
\noindent
\begin{minipage}[t]{0.48\linewidth}
\begin{verbatim}
START -> X1
X1 -> X2 X3
X2 -> PN
X2 -> D N
X3 -> V3 X1
\end{verbatim}
\end{minipage}\hfill
\begin{minipage}[t]{0.48\linewidth}
\begin{verbatim}
X3 -> V0
X3 -> V1 X2
X3 -> V2 X8
X8 -> P X2
\end{verbatim}
\end{minipage}
\begin{verbatim}
Count: 9
Grammar counts vector: [0, 0, 1, 2, 4], total strings: 7, Rule coverage length: 4
\end{verbatim}

\subsubsection*{English fragment with adjectival recursion and ambiguous PP adjunction (\(K=2\), global \#65)}
\begin{verbatim}
65. Global Optimum (Depth 10) Growth Rate: 2.671 [Indices: 201368]:
\end{verbatim}
\noindent
\begin{minipage}[t]{0.48\linewidth}
\begin{verbatim}
START -> X1
X1 -> X2 X3
X2 -> PN
X2 -> X2 X7
X2 -> D X5
X3 -> V3 X1
X3 -> X3 X7
\end{verbatim}
\end{minipage}\hfill
\begin{minipage}[t]{0.48\linewidth}
\begin{verbatim}
X3 -> V0
X3 -> V1 X2
X3 -> V2 X7
X5 -> A X5
X5 -> N
X7 -> P X2
\end{verbatim}
\end{minipage}
\begin{verbatim}
Count: 13
Grammar counts vector: [0, 0, 1, 2, 7], total strings: 10, Rule coverage length: 4
\end{verbatim}

\subsubsection*{SOV fragment with adjectival recursion (\(K=2\), global \#7)}
\begin{verbatim}
7. Global Optimum (Depth 8) Growth Rate: 1.852 [Indices: 2825]:
\end{verbatim}
\noindent
\begin{minipage}[t]{0.48\linewidth}
\begin{verbatim}
START -> X1
X1 -> X2 X3
X2 -> PN
X2 -> D X5
X3 -> X1 V3
X3 -> X2 V1
\end{verbatim}
\end{minipage}\hfill
\begin{minipage}[t]{0.48\linewidth}
\begin{verbatim}
X3 -> X9 V2
X3 -> V0
X5 -> A X5
X5 -> N
X9 -> X2 P
\end{verbatim}
\end{minipage}
\begin{verbatim}
Count: 11
Grammar counts vector: [0, 0, 1, 2, 5], total strings: 8, Rule coverage length: 4
\end{verbatim}

\subsection{Formal-Language Examples}

\subsubsection*{Dyck language with four bracket types (\(K=2\), global \#1)}
\begin{verbatim}
1. Global Optimum (Depth 10) Growth Rate: 5.151 [Indices: 19331]:
\end{verbatim}
\noindent
\begin{minipage}[t]{0.48\linewidth}
\begin{verbatim}
START -> X1
X1 -> X1 X1
X1 -> X2 RP4
X1 -> X4 RP3
X1 -> X6 RP2
X1 -> X8 RP1
X2 -> LP4
\end{verbatim}
\end{minipage}\hfill
\begin{minipage}[t]{0.48\linewidth}
\begin{verbatim}
X2 -> LP4 X1
X4 -> LP3
X4 -> X4 X1
X6 -> X6 X1
X6 -> LP2
X8 -> LP1
X8 -> LP1 X1
\end{verbatim}
\end{minipage}
\begin{verbatim}
Count: 14
Grammar counts vector: [0, 0, 4, 0, 32], total strings: 36, Rule coverage length: 4
START -> \epsilon
\end{verbatim}

\subsubsection*{General palindrome language (\(K=2\), global \#1)}
\begin{verbatim}
1. Global Optimum (Depth 8) Growth Rate: 1.767 [Indices: 62290]:
\end{verbatim}
\noindent
\begin{minipage}[t]{0.48\linewidth}
\begin{verbatim}
START -> X1
START -> X2
X1 -> X1 A
X1 -> A
X2 -> X4 B
X2 -> A X6
\end{verbatim}
\end{minipage}\hfill
\begin{minipage}[t]{0.48\linewidth}
\begin{verbatim}
X2 -> B
X4 -> B
X4 -> B X1
X4 -> B X2
X6 -> X2 A
\end{verbatim}
\end{minipage}
\begin{verbatim}
Count: 11
Grammar counts vector: [0, 2, 2, 4], total strings: 8, Rule coverage length: 3
START -> \epsilon
\end{verbatim}

\subsubsection*{Even-length palindrome language (\(K=1\), global \#1)}
\begin{verbatim}
1. Global Optimum (Depth 5) Growth Rate: 1.74 [Indices: 19]:
\end{verbatim}
\noindent
\begin{minipage}[t]{0.48\linewidth}
\begin{verbatim}
START -> X1
X1 -> X2 A
X1 -> X4 B
X2 -> A
\end{verbatim}
\end{minipage}\hfill
\begin{minipage}[t]{0.48\linewidth}
\begin{verbatim}
X2 -> A X1
X4 -> B
X4 -> B X1
\end{verbatim}
\end{minipage}
\begin{verbatim}
Count: 7
Grammar counts vector: [0, 0, 2, 0, 4], total strings: 6, Rule coverage length: 4
START -> \epsilon
\end{verbatim}

\subsubsection*{Odd-length palindrome language (\(K=2\), global \#1)}
\begin{verbatim}
1. Global Optimum (Depth 8) Growth Rate: 1.756 [Indices: 21796]:
\end{verbatim}
\noindent
\begin{minipage}[t]{0.48\linewidth}
\begin{verbatim}
START -> X2
START -> A
X2 -> X3 A
X2 -> X4 B
X2 -> B
\end{verbatim}
\end{minipage}\hfill
\begin{minipage}[t]{0.48\linewidth}
\begin{verbatim}
X3 -> A X2
X3 -> A A
X4 -> B X2
X4 -> B A
\end{verbatim}
\end{minipage}
\begin{verbatim}
Count: 9
Grammar counts vector: [0, 2, 0, 4], total strings: 6, Rule coverage length: 3
\end{verbatim}

\subsubsection*{\(a^n b^n\) language (\(K=1\), global \#1)}
\begin{verbatim}
1. Global Optimum (Depth 3) Growth Rate: 1.23 [Indices: 2]:
\end{verbatim}
\noindent
\begin{minipage}[t]{0.48\linewidth}
\begin{verbatim}
START -> X1
X1 -> X2 B
\end{verbatim}
\end{minipage}\hfill
\begin{minipage}[t]{0.48\linewidth}
\begin{verbatim}
X2 -> A
X2 -> A X1
\end{verbatim}
\end{minipage}
\begin{verbatim}
Count: 4
Grammar counts vector: [0, 0, 1, 0, 1], total strings: 2, Rule coverage length: 4
\end{verbatim}

\subsubsection*{Inherently ambiguous \(a,b,c\) union language (\(K=3\), global \#2)}
\begin{verbatim}
2. Global Optimum (Depth 9) Growth Rate: 1.271 [Indices: 7675344]:
\end{verbatim}
\noindent
\begin{minipage}[t]{0.48\linewidth}
\begin{verbatim}
START -> X1
X1 -> X2 X3
X1 -> X7 X8
X2 -> A X5
X3 -> X3 C
X3 -> C
X5 -> X2 B
\end{verbatim}
\end{minipage}\hfill
\begin{minipage}[t]{0.48\linewidth}
\begin{verbatim}
X5 -> B
X7 -> A
X7 -> A X7
X8 -> X9 C
X9 -> B
X9 -> B X8
\end{verbatim}
\end{minipage}
\begin{verbatim}
Count: 13
Grammar counts vector: [0, 0, 0, 1, 2, 4], total strings: 7, Rule coverage length: 5
\end{verbatim}

\section{Search Space and Guided Generation Exactness}
\label{app:canonical_guided}

\subsection{Canonical Adjacency}

Fix a finite indexed core rule universe
\[
\mathcal{R}_{\mathrm{core}}=\{r_1,\ldots,r_n\}
\]
with its fixed total order. The implementation uses variables
\(X_1,\ldots,X_M\). Since these names are arbitrary, the search is carried out
over a canonical naming convention rather than over all syntactic renamings of
the same grammar.

For a rule set \(C\), let \(I(C)\) be the set of variables
that occur in \(C\), and let
\[
h(C)=\max\{i:X_i\in I(C)\},
\]
with \(h(C)=0\) when \(I(C)=\emptyset\). A one-rule extension is
\emph{naming-admissible} if it respects the following convention:
\begin{enumerate}
    \item variables are introduced without gaps;
    \item a non-start variable may appear as the left-hand side of a binary rule
    only after it has already been introduced by a start projection or as a
    right-hand-side symbol; and
    \item if a binary rule introduces two fresh variables at once, they are
    named in left-to-right order as the next two available variables.
\end{enumerate}
Equivalently, at a node \(C\), the admissible additions are start
rules \(S\to X_i\) with \(i\le h(C)+1\), and binary rules
\(X_i\to X_jX_k\) whose left-hand side \(X_i\) is already in \(I(C)\), and whose
right-hand side uses existing variables, the next fresh variable, or the next two
fresh variables in left-to-right order. This is the implementation's
no-gap/no-orphan convention.

This convention does not remove any language-level solution. It only chooses one
representative from each class of grammars that differ by harmless renaming of variables.

\begin{proposition}[Canonical representatives]
\label{prop:canonical_representatives}
Let \(G\) be any reachable grammar in the normal-form
universe, ignoring the particular names of its variables. Then there
is a weakly equivalent representative \(G^{\mathrm{can}}\) whose structural
rules satisfy the naming-admissibility convention above.
\end{proposition}

\begin{proof}
First remove any unreachable rules from \(G\). This does not change
the language generated from \(S\). The remaining variables are all
reachable from the start symbol.

We now rename the reachable variables in order of first introduction.
Start with no variables named. Whenever a start projection or a rule
whose left-hand side has already been named first exposes an unnamed structural
variable, assign that variable the next unused name \(X_i\). If two unnamed
variables are first exposed as the two children of the same binary rule, assign
the left child the next name and the right child the following name. Continue
until all reachable variables have been named.

The resulting rule set uses a prefix \(X_1,\ldots,X_t\) of the variable names, so there are no gaps. Moreover, every non-start left-hand side
\(X_i\) has already been introduced before rules with left-hand side \(X_i\) are
expanded, because \(X_i\) received its name when it first became reachable from
a start projection or from the right-hand side of an already named parent. Thus
the renamed rule set has no orphaned binary-rule left-hand sides. The renaming is a bijection on variables. It leaves the start symbol fixed and
transports the preterminal interface along the same renaming while keeping
preterminal labels fixed. Therefore it preserves the generated preterminal
language. Therefore \(G^{\mathrm{can}}\) is weakly equivalent to \(G\) and is
naming-admissible.
\end{proof}

In an inner search rooted at a rule set \(R_0\), each node stores the largest
index \(m\) added on the path from \(R_0\), with \(m=0\) at the root. The
canonical adjacents of a node \(C\) are
\[
\operatorname{Adj}(C,m)=
\{(C\cup\{r_j\},j):j>m,\ r_j\notin C,\ C\cup\{r_j\}
\text{ is naming-admissible from }C\}.
\]
Thus canonical adjacency combines two restrictions: the naming-admissibility
gates above, and the strictly increasing rule-index order used to remove
permutation duplicates.

Let \(\mathcal{A}(R_0)\) denote the admissible generated universe from root
\(R_0\): the rule sets \(C^*\supseteq R_0\) whose added rules can be listed in
increasing fixed rule order while every prefix remains
naming-admissible.

\begin{theorem}[Canonical adjacency spans the admissible rule-superset space exactly once]
\label{thm:canonical_adjacency}
For every \(C^*\in\mathcal{A}(R_0)\), there is exactly one canonical path from
\(R_0\) to \(C^*\). Consequently, canonical adjacency generates every
admissible rule set once, without duplicate paths induced by
rule-addition permutations. By Proposition~\ref{prop:canonical_representatives},
this does not lose language-level solutions, modulo renaming of structural
nonterminals.
\end{theorem}

\begin{proof}
Let
\[
C^*\setminus R_0=\{r_{j_1},\ldots,r_{j_k}\}
\]
with \(j_1<\cdots<j_k\). Since \(C^*\in\mathcal{A}(R_0)\), adding these rules
in increasing order preserves the naming-admissibility convention at
each prefix. At the first step, \(j_1>0\), so \(r_{j_1}\) is permitted by the
root convention. At step \(\ell>1\), the stored index is \(m=j_{\ell-1}\), and
\(j_\ell>j_{\ell-1}\). Therefore \(r_{j_\ell}\) satisfies the canonical
index constraint. Because the prefixes are naming-admissible by membership in
\(\mathcal{A}(R_0)\), each step is an element of
\(\operatorname{Adj}(C,m)\). Hence a canonical path from \(R_0\) to \(C^*\)
exists.

For uniqueness, consider any canonical path from \(R_0\) to \(C^*\). The path
must add exactly the rules in \(C^*\setminus R_0\). The canonical index
constraint requires the sequence of added rule indices to be strictly
increasing. But the only strictly increasing ordering of the set
\(\{j_1,\ldots,j_k\}\) is
\[
(j_1,\ldots,j_k).
\]
Thus every canonical path to \(C^*\) uses the same sequence of rule additions,
and the path is unique.
\end{proof}

\subsection{Exactness of Wildcard-Guided Generation}

Canonical adjacency defines the candidate universe, while guided generation is
a filter over that universe. Fix a root rule set \(R\), a current
inner-search node
\[
C=R\cup\Delta,
\]
and an ordered canonical candidate list
\[
\operatorname{Cand}(C)=\{c_1,\ldots,c_q\}.
\]
For a structural-preterminal mapping \(\sigma_j\) associated with the root, let
\(u_j\) be the first shortest evidence string not parsed under that mapping.
Write
\[
R_{\mathrm{pre}}(\sigma_j)=\{A\to \sigma_j(A):A\in\operatorname{dom}(\sigma_j)\}.
\]

For a candidate \(c_t\), the guide considers the concrete rules
\[
R\cup\Delta\cup\{c_t\}\cup R_{\mathrm{pre}}(\sigma_j).
\]
It then applies a wildcard relaxation: for this candidate and mapping,
each non-start nonterminal that is currently unproductive in the concrete
grammar may stand for one nonempty span of \(u_j\). Productivity is evaluated
candidate-wise; if \(c_t\) itself makes a nonterminal productive, that
nonterminal is not treated as a wildcard for bit \(t\).

\begin{definition}[Wildcard survivor predicate]
\label{def:wildcard_survivor}
Candidate \(c_t\) survives mapping \(j\), written
\(\operatorname{Surv}_j(c_t)\), if there is a wildcard-relaxed derivation from
\(S\) matching \(u_j\) that uses every rule in \(\Delta\) at least once and
uses \(c_t\) at least once.
\end{definition}

\begin{theorem}[Exactness of successful guided generation]
\label{thm:guide_exactness}
If the guided bitset dynamic program succeeds at node \(C=R\cup\Delta\), the
candidates it enqueues are exactly
\[
\operatorname{Guide}(C)=
\{c_t\in\operatorname{Cand}(C):\exists j\ \operatorname{Surv}_j(c_t)\}.
\]
If the guided filter cannot be applied safely, the implementation enqueues the
entire canonical list \(\operatorname{Cand}(C)\). Therefore guided generation
does not remove any candidate satisfying the wildcard survivor predicate.
\end{theorem}

\begin{proof}
Fix one mapping \(j\). The guide maintains two span tables
\[
\operatorname{NoUse}[A,M,i,h],
\qquad
\operatorname{Used}[A,M,i,h],
\]
whose entries are bitsets over the candidate positions. For candidate bit \(t\),
membership in \(\operatorname{NoUse}[A,M,i,h]\) means that, using the wildcard
relaxation for \(R\cup\Delta\cup\{c_t\}\cup R_{\mathrm{pre}}(\sigma_j)\),
nonterminal \(A\) derives a pattern matching the slice \(u_j[i:h]\), uses
exactly the subset \(M\) of delta rules, and has not used \(c_t\). Membership
in \(\operatorname{Used}[A,M,i,h]\) has the same meaning except that the
derivation has used \(c_t\) at least once. The mask records rule occurrence as
a set, so repeated uses of a delta rule do not change the mask.

We prove this invariant by induction on span length, with a monotone closure
for same-span start projections. Preterminal interface rules in
\(R_{\mathrm{pre}}(\sigma_j)\) initialize exactly the length-one spans whose
observed preterminal matches the corresponding symbol of \(u_j\). Wildcard
leaves initialize exactly the positive-length spans for candidate-wise
unproductive non-start nonterminals. These entries belong to
\(\operatorname{NoUse}\), since neither initialization has used \(c_t\).

For start projections \(S\to A\), the contents of the child cell for \(A\) are
copied to the corresponding parent cell for \(S\). If the projection belongs to
\(\Delta\), its delta bit is added to the mask. If the projection is the
candidate \(c_t\), the result is written to \(\operatorname{Used}\), because
the candidate has now been used. The implementation applies these transfers as
a monotone closure on each span; since the searched normal form has no arbitrary unit-rule cycles, this closure computes exactly the least set of
same-span projection consequences.

For a binary rule \(A\to BC\), split point \(s\), and child masks \(M_1,M_2\),
the parent mask is
\[
M_1\cup M_2\cup M_r,
\]
where \(M_r\) is the delta bit if the applied rule lies in \(\Delta\), and
\(0\) for root rules or the candidate rule \(c_t\); use of \(c_t\) is tracked
separately by the \(\operatorname{Used}\) table. Candidate bit
\(t\) is propagated only when both child derivations are valid for the same
candidate \(c_t\). If neither child has used \(c_t\), the result is written to
\(\operatorname{NoUse}\); if at least one child has used \(c_t\), the result is
written to \(\operatorname{Used}\). When the applied binary rule itself is
\(c_t\), the result is written directly to \(\operatorname{Used}\).

These transitions are exactly the productions of the candidate-specific
wildcard grammar, together with the bookkeeping for delta-rule masks and
candidate use. Hence bit \(t\) belongs to
\[
\operatorname{Used}[S,M_\Delta,0,|u_j|],
\]
where \(M_\Delta\) is the full mask over \(\Delta\), if and only if
\(\operatorname{Surv}_j(c_t)\) holds.

The implementation unions these final bitsets over all active mappings \(j\).
Therefore, when the guided dynamic program succeeds, the enqueued candidates
are precisely those in \(\operatorname{Guide}(C)\).If the program cannot complete the guided pass, the caller enqueues all
canonical adjacents, which removes no survivor.
\end{proof}

The theorem is prefix-local. At node \(C=R\cup\Delta\), survival of \(c_t\) is
tested using only the root \(R\), the already-added prefix \(\Delta\), the
single candidate \(c_t\), and the active mappings \(\sigma_j\). Later canonical
rules are not available to the guide. Thus the guide does not impose a
top-down parse-tree order: a surviving candidate may occur above, below, or
beside rules in \(\Delta\) in the wildcard derivation. The order is only the
search-prefix order. Completeness of a guided run therefore assumes that the
needed canonical increment has a path whose successive prefixes satisfy this
survivor predicate, unless the guide falls back to the full canonical list.

\vskip 0.2in
\bibliography{GreedyGrammarInduction}

\end{document}